\documentclass[11pt]{article}
\usepackage{eacl2017}

\usepackage{times}
\usepackage{latexsym}
\usepackage{url}
\usepackage{arydshln}
\usepackage{graphicx}
\usepackage[fleqn]{amsmath}
\usepackage{amsfonts, amssymb}
\usepackage{xcolor}
\usepackage{multirow}
\usepackage{breqn}

\newcommand{\todooff}{\long\gdef\todo##1##2{}}
\newcommand{\todoon}{\long\gdef\todo##1##2{{
\bf\textcolor{red} {TODO: ##1: ##2}
}}}
\todoon
\todooff

\newcounter{notecounter}
\newcommand{\enotesoff}{\long\gdef\enote##1##2{}}
\newcommand{\enoteson}{\long\gdef\enote##1##2{{
\stepcounter{notecounter}
{\large\bf
\hspace{1cm}\arabic{notecounter} $<<<$ ##1: ##2
$>>>$\hspace{1cm}}}}}
\enoteson
\enotesoff

\def\dnrm#1{\mbox{$_{\hbox{\scriptsize #1}}$}}

\def\figref#1{Figure~\ref{fig:#1}}
\def\figlabel#1{\label{fig:#1}\label{p:#1}}

\def\tabref#1{Table~\ref{tab:#1}}
\def\tablabel#1{\label{tab:#1}\label{p:#1}}

\def\secref#1{Section~\ref{sec:#1}}
\def\seclabel#1{\label{sec:#1}\label{p:#1}}
\def\eqref#1{Eq.~\ref{eqn:#1}}
\def\eqlabel#1{\label{eqn:#1}}

\def\pc2t{$P(t|c)$}
\def\pc2ti#1{$P(t_#1|c)$}

\eaclfinalcopy
\def\eaclpaperid{144}

\title{Noise Mitigation  for Neural Entity Typing and Relation Extraction}

\author{Yadollah Yaghoobzadeh* \and Heike Adel* \and Hinrich Sch\"{u}tze\\
* \textit{These authors contributed equally to this work}\\
	    Center for Information and Language Processing\\
	    LMU Munich, 
	    Germany\\
	    {\tt {yadollah|heike}@cis.lmu.de}}

\date{}

\begin{document}

\maketitle

\begin{abstract}
In this paper, we address two different types of noise 
in information extraction models:
noise from distant supervision
and noise from pipeline input features. 
Our target tasks are entity typing
and relation extraction. 
For the first noise type, we introduce 
multi-instance multi-label learning algorithms 
using neural network models,
and apply them to 
fine-grained entity typing for the first time. 
Our model outperforms the state-of-the-art 
supervised approach which uses global embeddings of entities.
For the second noise type, 
we propose 
ways to improve the integration of 
noisy entity type predictions into relation extraction. 
Our experiments show that  probabilistic predictions 
are more robust than discrete predictions
and that joint training of the two tasks performs best.
\end{abstract}

\section{Introduction}

Knowledge bases (KBs) are important resources for 
natural language processing
tasks
like
question answering 
and
entity linking.
However, KBs are far from complete (e.g., \newcite{socher2013reasoning}).
Therefore, methods for automatic knowledge base completion (KBC)
are beneficial.
Two subtasks of KBC are 
\emph{entity typing (ET)}
and 
\emph{relation extraction (RE)}. 
We address both tasks in this paper.

As in other information extraction tasks,
obtaining labeled training data for ET and RE is challenging. The
challenge grows as
labels become more fine-grained.
Therefore, distant supervision \cite{Mintz09} is widely used.
It reduces the need for manually created resources.
Distant supervision assumes that 
if 
an entity has a type (resp.\ two entities have a relationship)
in a KB, 
then all sentences mentioning 
that entity (resp.\ those two entities)
express that type (resp.\ that relationship).
However, that
assumption is too
strong and gives rise to many \emph{noisy} labels.
Different techniques
to deal with that problem 
have been investigated.
The main technique  is
multi-instance (MI) learning \cite{riedel2010}. 
It relaxes the distant supervision assumption to the assumption
that at least one instance of a bag (collection of 
all sentences containing the given entity/entity pair) 
expresses the type/relationship given in the KB.
Multi-instance multi-label (MIML) learning 
is a generalization of MI in which one bag
can have several labels \cite{surdeanu2012multi}. 

Most MI and MIML methods are based on hand crafted features.
Recently, \newcite{Zeng15emnlp} introduced an end-to-end
approach to
MI learning based on neural networks.
Their MI method takes the most confident instance
as the prediction of the bag.
\newcite{lin2016} further improved that method  by taking
other instances into account as well;
they proposed MI learning based on
selective attention as an alternative way of
relaxing the impact of noisy labels on RE. 
In selective attention, a weighted average of instance
representations is calculated first and then used 
to compute the prediction of a bag.

In this paper, we introduce two multi-label versions of 
MI.
(i) \emph{MIML-MAX}
takes the maximum instance
for each label. (ii) \emph{MIML-ATT} applies, for each label, selective attention 
to the instances.
We apply MIML-MAX and MIML-ATT to fine-grained ET.
In contrast to RE, the ET
task we consider contains a larger
set of labels, with a variety of different 
granularities and hierarchical relationships. 
We show that 
MIML-ATT 
deals well with noise in
corpus-level ET and 
improves or matches the results of a supervised
model based on global embeddings of entities.

The second type of noise we address in this paper influences 
the integration 
of ET into RE.
It has been shown that adding entity types as features improves 
RE models (cf.\ \newcite{ling2012fine}, \newcite{liu}).
However, noisy training data and difficulties of
classification often cause
wrong predictions of ET and,
as a result, noisy inputs to RE. 
To address this, we propose a 
joint model of ET and RE and
 compare it
with methods that 
integrate
ET results in a strict pipeline.
The joint model performs best. Among 
the pipeline models, we show 
that using probabilities instead of binary decisions 
better deals with noise (i.e., possible ET errors).

To sum up, our contributions are as follows. 
(i) We introduce new algorithms for MIML using neural networks.
(ii) We apply MIML to fine-grained entity typing for the first time
 and show that it outperforms the state-of-the-art supervised method based on entity embeddings.
(iii) We show that a novel way of integrating noisy
entity type predictions into a relation extraction model
and joint training of the two tasks lead to large
improvements of RE performance.

We release code and data for future research.\footnote{\url{cistern.cis.lmu.de}}

\section{Related Work}
\seclabel{related}
\textbf{Noise mitigation for distant supervision.}
Distant supervision can be used to train
information extraction systems, e.g., in 
relation extraction (e.g., \newcite{Mintz09}, \newcite{riedel2010},
\newcite{hoffmann2011}, \newcite{Zeng15emnlp})
and entity typing (e.g.,
\newcite{ling2012fine}, \newcite{yogatama2015acl}, \newcite{donghybrid}).
To mitigate the noisy label problem, 
multi-instance (MI) learning has been introduced
and applied in relation extraction 
\cite{riedel2010,ritter13tacl}. 
\newcite{surdeanu2012multi} introduced multi-instance multi-label (MIML) learning
to extend MI learning for multi-label relation extraction.
Those models are based on manually designed features.
\newcite{Zeng15emnlp} and \newcite{lin2016} introduced MI 
learning methods for neural networks. 
We introduce MIML algorithms
for neural networks.
In contrast to most MI/MIML methods, which are applied in relation extraction, 
we apply MIML to the task of 
fine-grained entity typing.
\newcite{ritter13tacl} applied MI 
on a Twitter dataset with ten types.
Our dataset has a larger number of classes or types (namely 102)
and input examples, 
compared to that Twitter dataset and also to
the most widely used datasets for evaluating MI (cf.\ \newcite{riedel2010}).
This makes our setup more challenging 
because of different dependencies
and the multi-label nature of the problem. 
Also, there seems to be a difference between how entity relations and
entity types
are expressed in text.
Our experiments support that hypothesis.

\textbf{Knowledge base completion (KBC).}
Most KBC systems focus on identifying triples $R(e_1,r,e_2)$
missing from a KB \cite{nickel12yago,bordes2013transe,Weston2013Conn,socher2013reasoning,Jiang12linkpred,Riedel13universal,Wang14joint}. 
Work on entity typing or unary relations for
KBC is more recent \cite{Yao2013,neelakantan2015inferring,figment15,figment17}.
In this paper, we build a KBC system for unary and binary relations
using contextual information of words and entities.

\textbf{Named entity recognition (NER) and typing.}
NER systems (e.g.,
\newcite{finkel2005}, \newcite{cw11})
used to consider only a small set of entity types. Recent work
also addresses fine-grained NER
\cite{spaniol2012hyena,ling2012fine,yogatama2015acl,donghybrid,delcorro15finet,xiang2016afet,partialLabel16,attentiveTyper16}.  
Some of this work (cf. \newcite{yogatama2015acl}, \newcite{donghybrid})
treats entity
segment boundaries  as given and 
classifies mentions into fine-grained types.
We make a similar assumption, but in contrast to NER,
we evaluate  on the corpus-level entity typing task
of \newcite{figment15}; thus, we do not
need test sentences annotated with context
dependent entity types.
This task was also used to evaluate  embedding learning methods
\cite{derata2016}. 

\textbf{Entity types for relation extraction.}
Several studies have integrated entity
type information into relation extraction -- either
coarse-grained \cite{hoffmann2011,zhou} or fine-grained
\cite{liu,du,augenstein,vlachos,Yao2010joint,ling2012fine} entity types.
In contrast to most of this work, but similar to \newcite{Yao2010joint}, we do not incorporate binary entity type values,
but probabilistic outputs.  
Thus, we allow the relation extraction system
to compensate for errors of entity typing.
Additionally, we compare this approach to various other possibilities,
to investigate which approach performs best.
 \newcite{Yao2010joint} found that joint training of
 entity typing and relation extraction is better than a pipeline 
 model; we show that this result also holds for
neural network models and when the  number of entity types is large.

\section{MIML Learning for Entity Typing}
\seclabel{main}
Entity typing (ET) is the task of finding, for each 
named entity, a set of types or classes 
that it belongs to, e.g., ``author''
and ``politician''
for ``Obama''.
Our goal is corpus-level prediction of entity types.
We use the  entity-type information
from a KB and 
annotated contexts of entities in
a corpus 
\enote{ha}{maybe better: ... from a KB and a
corpus annotated with entities}
to estimate $P(t|e)$,
the probability that entity $e$ has
type $t$.

More specifically, consider an entity $e$
and 
$B = \{c_1 ,c_2 , ... ,c_{q}\}$, the set of $q$
contexts of $e$ in the corpus.
Each $c_i$ is an instance of $e$ and since $e$ can have
several labels, it is a multi-instance multi-label (MIML) learning
problem.
We address MIML using neural networks 
by representing each context as a vector $\vec{c_i} \in \mathbb{R}^{h}$,
and learn $P(t|e)$ from the set of contexts of entity $e$.
In the following, we first describe our MIML algorithms 
and then explain how $\vec{c_i}$ is computed.

\textbf{Notations and definitions.}
Lowercase letters (e.g., $e$) refer to variables.
Lowercase letters with an upper arrow (e.g., $\vec{e}$) are vectors.
We define $\mbox{BCE}$, binary cross entropy,
as follows where $y$ is a binary variable and 
$\hat{y}$ is a real valued variable between 0 and 1.
\begin{equation}
\mbox{BCE}(y, \hat{y}) = - \Big(y \log(\hat{y}) + (1 - y) (1 - \log(\hat{y})) \Big)\nonumber
\end{equation}

\subsection{Algorithms}
\textbf{Distant supervision.}
The basic way to estimate $P(t|e)$ is based on distant supervision 
with learning the type probability of each $c_i$ individually,
by making the assumption that each $c_i$ expresses all 
labels of $e$.
Therefore, we define the context-level probability
function as:
\begin{equation}
P(t|c_i) = \sigma (\vec{w_t} \vec{c_i} + b_t)
\eqlabel{cm}
\end{equation}
where $\vec{w_t} \in \mathbb{R}^{h}$ is the output weight vector and 
$b_t $ is the bias scalar for type $t$.
The cost function is defined based on binary cross entropy:
\begin{dmath}
L(\theta) = 
\sum_{c} \sum_{t}{\mbox{BCE}(y_t, P(t|c)})
\eqlabel{loss} 
\end{dmath}
where  $y_t$ is 1 if entity $e$ has type $t$
otherwise 0.
To compute  $P(t|e)$ at prediction time, i.e., $P\dnrm{pred}(t|e)$,
the context-level probabilities must be aggregated. 
Average is the usual way of doing that:
\begin{equation}
P\dnrm{pred}(t|e) = \dfrac{1}{q} \sum_{i = 1}^{q} P(t|c_i)
\end{equation}

\textbf{Multi-instance multi-label.}
The distant supervision assumption is
that \emph{all} contexts of an entity
with type $t$ are contexts of $t$; e.g.,
we label all contexts mentioning
``Barack Obama'' with all of his types.
Obviously, the labels
are incorrect or \emph{noisy} for some contexts.
Multi-instance multi-label (MIML) learning addresses this problem.
We apply MIML to fine-grained ET  for the first time.
Our assumption is: if entity $e$ has type $t$, then
there is at least one  context of $e$ in the corpus in which
$e$ occurs as type $t$.
So, we apply this assumption during training
with the following estimation of the type probability of an entity:
\begin{equation}
P(t|e) = \max_{1 \le i \le q} P(t|c_i)
\eqlabel{miml}
\end{equation}
which means we take the \textbf{maximum} probability of type $t$
over all contexts of entity $e$ as $P(t|e)$.
We call this approach \textbf{MIML-MAX}.

MIML-MAX picks the most confident context for $t$,
ignoring the probabilities of all the other contexts. 
Apart from missing information, this can be especially harmful if the entity annotations in the corpus are
the result of an entity linking system. 
In that case, 
the most confident context might be wrongly linked to the entity.
So, it can be beneficial to leverage all contexts into the final prediction,
e.g., by \textbf{averaging} the type probabilities
of all contexts of entity $e$:
\begin{equation}
P(t|e) = \dfrac{1}{q} \sum_{i = 1}^{q} P(t|c_i)
\eqlabel{avg_inst}
\end{equation}
We call this approach \textbf{MIML-AVG}.
We also propose a combination of the maximum and average,
which uses MIML-MAX (\eqref{miml}) in training 
and MIML-AVG (\eqref{avg_inst}) in prediction.
We call this approach \textbf{MIML-MAX-AVG}.

MIML-AVG treats every context equally which might
be problematic since many contexts are irrelevant 
for a particular type.
A better way is to weight the contexts according to their
similarity to the types.
Therefore, we propose using selective \textbf{attention}
over contexts as follows and call this approach \textbf{MIML-ATT}.
MIML-ATT is the multi-label version of the selective attention method 
proposed in \newcite{lin2016}. 
To compute the type probability for $e$, we define:
\begin{equation}
P(t|e) = \sigma (\vec{w_{t}} \vec{a_t} + b_t) 
\end{equation}
where $\vec{w_t} \in \mathbb{R}^{h}$ is the output weight vector and 
$b_t $ the bias scalar for type $t$, and 
$\vec{a_t}$ is the aggregated representation of all contexts $c_i$
of $e$ for type $t$, computed as follows:
\begin{equation}
\vec{a_t} = \sum_{i}{ \alpha_{i,t} \vec{c_i}}
\end{equation}
where $\alpha_{i,t}$ is the attention score of context $c_i$ for type $t$
and
$\vec{a_t} \in \mathbb{R}^{h}$ can be interpreted as the representation of entity $e$
for type $t$.

$\alpha_{i,t}$ is defined as:
\begin{equation}
\alpha_{i,t} = \dfrac{\exp(\vec{c_i} \textbf{M} \vec{t})}
{\sum_{j=1}^{q}{\exp(\vec{c_j} \textbf{M} \vec{t}})}
\eqlabel{alpha}
\end{equation}
where $\textbf{M} \in \mathbb{R} ^ {h \times d_t}$ is a weight matrix that 
measures the similarity of $\vec{c}$ and $\vec{t}$. 
$\vec{t} \in \mathbb{R}^{d_t}$ 
is the representation of type $t$.

\tabref{mimlmodels} summarizes the differences
of our MIML methods with respect to the aggregation function they use
to get corpus-level probabilities.
For optimization of all MIML methods, we use the binary cross entropy loss function,  
\begin{equation}
L(\theta) = \sum_{e} \sum_{t} \mbox{BCE}(y_t, P(t|e))
\eqlabel{miml_loss} 
\end{equation}
In contrast to the loss function of distant supervision in \eqref{loss}, 
which iterates over all \emph{contexts},
we iterate over all \emph{entities} here.

\begin{table}[t]
\begin{center}
\footnotesize
\begin{tabular}{l|c|c}
\multicolumn{1}{c}{Model} & Train & Prediction \\ 
\hline 
MIML-MAX & MAX & MAX \\ 
\hline 
MIML-AVG & AVG & AVG \\ 
\hline 
MIML-MAX-AVG & MAX & AVG \\ 
\hline 
MIML-ATT & ATT & ATT
\end{tabular} 
\caption{Different MIML algorithms for entity typing, and the aggregation
function they use to get corpus-level  probabilities.}
\tablabel{mimlmodels}
\end{center}
\end{table}

\subsection{Context Representation}
\seclabel{model}
To produce a high-quality
context representation $\vec{c}$, 
we use convolutional neural networks (CNNs).

The first layer of the CNN is a
\emph{lookup table} that maps each word in $c$ to an 
embedding of size $d$. 
The output of the lookup layer 
is a matrix $E \in \mathbb{R}^{d
  \times s}$ (the embedding layer), 
  where $s$ is the context size (a fixed number of words).

The CNN
uses $n$ filters of different window widths $w$
to narrowly convolve $E$.
For each of the $n$ filters $H \in \mathbb{R}^{d\times w}$,
the result of applying $H$ to  matrix $E$
is a feature map $\vec{m} \in \mathbb{R}^{s-w+1}$:
\begin{equation}
\mbox{m}[i] = g(E_{:, i : i + w - 1} \odot H)
\end{equation}
where 
$g$ is the $relu$ function,
$\odot$ is the Frobenius product,
$E_{:, i : i + w - 1}$ are the columns $i$ to $i + w - 1$
of $E$ and
 $ 1\leq w\leq k$ are the window widths we consider.
Max pooling then gives us one feature for each
filter and the concatenation of those features 
is the CNN representation of $c$.

As it is shown in the entity typing part of \figref{combinedCNNs},
we apply the CNN to the left and right context of the entity mention
and the concatenation $\vec{\phi}(c) \in \mathbb{R}^{2n}$
is fed into a multi-layer perceptron (MLP)
to get the final context representation $\vec{c} \in \mathbb{R}^{h}$:
\begin{equation}
\vec{c} = \tanh\Big(\textbf{W}_h \vec{\phi}(c)\Big)
\end{equation}

\begin{figure*}
\centering
\includegraphics[width=340pt,height=170pt]{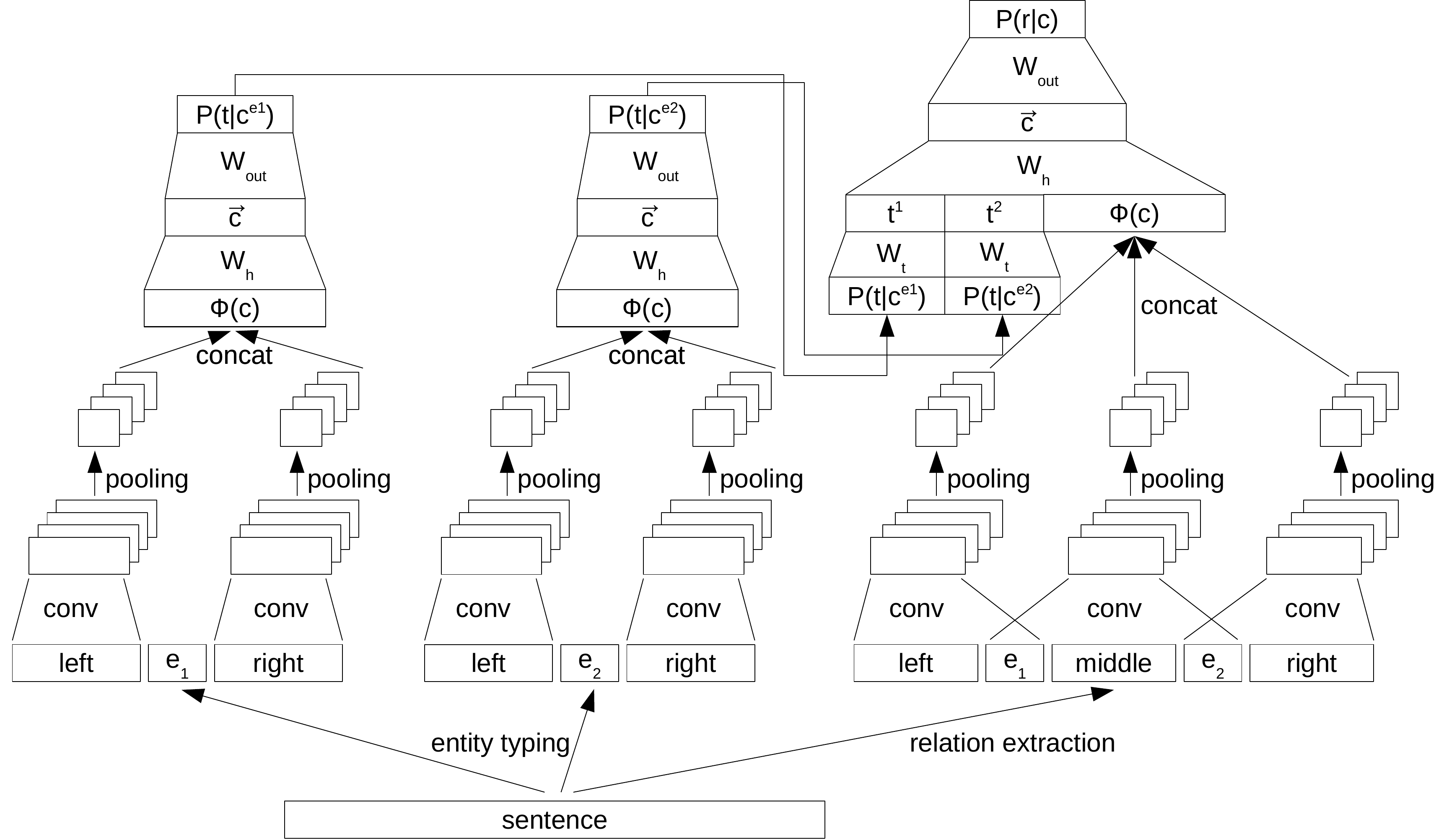}
\caption{Our architecture for joint entity typing and relation extraction}
\figlabel{combinedCNNs}
\end{figure*}

\section{Type-aware Relation Extraction}
Relation extraction (RE) is mostly defined as  finding relations between
pairs of entities, for instance, finding the relation ``president-of'' 
between  ``Obama'' and ``USA''.
Given a set of $q$ contexts for an entity pair $z$,
$B = \{c_1 ,c_2 , ... ,c_{q}\}$
in the corpus,
we learn  $P(r|z)$, which is the probability of relation $r$ for $z$.
We assume that each $z$ has one relation $r(z)$.
Each $c_i$ is represented by a vector $\vec{c_i} \in \mathbb{R}^{h}$,
which is our type-aware representation of context described
in \secref{contextrepresentation}.

To learn $P(r|z)$, we use the multi-instance (MI) learning 
method of \newcite{Zeng15emnlp}:
\begin{equation}
\left.\begin{aligned}
&P(r|c_i) =
  \mbox{softmax}
  \Big(\textbf{W}\dnrm{out} \vec{c_i} \Big), \\
&P(r|z) = \max_{1 \leq i \leq q} P(r|c_i)   
\end{aligned} \right.
\eqlabel{reProb}
\end{equation}
where $P(r|c_i)$ is  the probability of relation $r$ for context $c_i$.
The cost function we optimize is:
\begin{equation*}
L(\theta) = - \sum_{z}
{\log{P(r(z)|z)}} \eqlabel{loss_rel}
\end{equation*}

\subsection{Context Representation}
\seclabel{contextrepresentation}
Similar to our entity typing system, we apply CNNs
to compute the context representation $\vec{\phi}(c)$.
In particular, we 
use \newcite{adel2016}'s CNN. 
It uses an input representation designed for RE.
Each sentence is split into three parts:
left of the relation arguments, between the relation arguments and
right of the relation arguments. The parts ``overlap'',
i.e., the left (resp.\ right) argument is included in both 
left (resp.\ right) and middle parts.
For each of the three parts,
convolution and 3-max pooling \cite{kalchbrenner2014} is performed. The context
representation $\vec{\phi}(c) \in \mathbb{R}^{3\cdot 3\cdot n}$ is the concatenation of the pooling results.

\subsubsection{Integration of Entity Types}
We concatenate the entity type 
representations  $\vec{t^1} \in \mathbb{R}^{\tau}$ and $\vec{t^2}\in \mathbb{R}^{\tau}$ of 
the relation arguments
to the CNN representation of the context, $\vec{\phi}(c)$:
\begin{equation}
\vec{\phi}(c)' = [\vec{\phi}(c):\vec{t^1}:\vec{t^2}]
\label{eq:concat}
\end{equation}
Our context representation $\vec{c}$ is then:
\begin{equation}
\vec{c} = \tanh\Big(\textbf{W}_h \vec{\phi}(c)'\Big)
\end{equation}
where $\textbf{W}_h \in \mathbb{R} ^{h \times (3 \cdot 3 \cdot n+2\tau)}$
is the weight matrix.
This is also
depicted in Figure \ref{fig:combinedCNNs}, right column,
third layer from the top: $\vec{t^1}$, $\vec{t^2}$, $\vec{\Phi}(c)$.
We calculate $\vec{t^1}$ and $\vec{t^2}$ from the predictions
of the entity typing model with the following transformation: 
\begin{equation}
\vec{t^k} = f\Big(\textbf{W\dnrm{t}}
 \big[ P(t_1|c^{e_k}) \ldots P(t_T|c^{e_k}) \big]\Big)
\eqlabel{hidden}
\end{equation}
where $c^{e_k}$ is the context of $e_k$, 
$\textbf{W\dnrm{t}}\in \mathbb{R}^{{\tau}\times T}$ is a weight
matrix (learned from corpus or during training) and $f$ is a function (identity or $\tanh$). 
With the transformation $\textbf{W\dnrm{t}}$, 
the model can combine predictions for
different types to learn better internal representations 
$t^1$ and $t^2$.
The choices of $\textbf{W\dnrm{t}}$ and $f$ depend on the 
different representations we investigate and describe
in the following.

\textbf{(1) Pipeline:}
We integrate entity types into the RE model, using the output
of ET in a pipeline model (see \eqref{hidden}).
We test the following representations of $\vec{t^k}$, $k \in \{1,2\}$.
\textbf{PREDICTED-HIDDEN}: $\textbf{W\dnrm{t}}$ 
from \eqref{hidden} is learned during training and $f$ is tanh.
That means that a hidden layer learns representations based on the predictions
$P(t_1|c^{e_k}) \ldots
P(t_T|c^{e_k})$.
\textbf{BINARY-HIDDEN}: This is the binarization of the input of 
PREDICTED-HIDDEN, i.e., each 
probability estimate is converted to 0 or 1 (with a threshold of 0.5).
\textbf{BINARY}: $\vec{t^k}$ is the binary vector itself (used by \newcite{ling2012fine}).
\textbf{WEIGHTED}: The columns of matrix $\textbf{W\dnrm{t}}$ 
from \eqref{hidden}
are the distributional embeddings of types trained 
on the corpus (see Section \ref{sec:embeddings}).
$f$ is the identity function.

\textbf{(2) Joint model:}
As an alternative to the pipeline model, 
we investigate integrating entity typing into RE by jointly
training both models. 
We use the architecture depicted in \figref{combinedCNNs}.
The key difference to the pipeline model PREDICTED-HIDDEN 
is that 
we  learn $P(t|c)$ and $P(r|c)$ jointly, called \textbf{JOINT-TRAIN}.
We compare JOINT-TRAIN
to other models, including the pipeline models.

During training of JOINT-TRAIN, we compute the cost of the ET
model for typing the first entity $L_1(\theta_T)$, the cost
for typing the second entity $L_2(\theta_T)$ and
the cost of the RE model 
for assigning a relation to the two entities $L(\theta_R)$.
Then, we combine those costs with a weight $\gamma$ 
which is tuned on the development set:
\begin{equation*}\label{jointcost}
\left.\begin{aligned}
&L(\theta) = \sum_{z}{\Big(L_1(\theta_T) + L_2(\theta_T) + \gamma \cdot
L(\theta_R)}\Big), \\
&L_i(\theta_T) = \sum_{t}{\mbox{BCE}(y^{e_i}_t, P(t|c^{e_i}))}, \\
&L(\theta_R) = - \log{P(r(z)|z)}
\end{aligned}\right.
\end{equation*}
$P(r|z)$ is computed based on \eqref{reProb}.

Note that based on this equation, 
the ET parameters are optimized 
on the contexts of the RE examples, which are 
a subset 
of all training examples of ET.
However in the pipeline models, ET 
is trained on the whole training set used for typing.
Also note that in JOINT-TRAIN we do not use MIML 
for the ET part but a distant supervised cost function.

\section{Experimental Data, Setup and Results}
\seclabel{data}
For entity typing, we use
CF-FIGMENT \cite{figment16url},
a dataset published by \newcite{figment15}.  CF-FIGMENT is
derived from a version of ClueWeb \cite{clueweb16url} in which Freebase entities
are annotated using
FACC1
\cite{lemur16url,gabrilovich2013facc1}.  CF-FIGMENT contains 200,000
Freebase entities that were mapped to 102 FIGER types
\cite{ling2012fine}, divided into train (50\%), dev (20\%) and test (30\%); and a
set of 4,300,000 sentences (contexts) containing those entities.

\begin{table}
\footnotesize
 \begin{tabular}{l@{\hspace{0.2cm}}l}
  GOV.GOV\_agency.jurisdiction &   PPL.PER.children\\
  GOV.us\_president.vice\_president&  PPL.PER.nationality\\
  PPL.deceased\_PER.place\_of\_death&  PPL.PER.religion\\
  ORG.ORG.place\_founded & 
  PPL.PER.place\_of\_birth\\
  ORG.ORG\_founder.ORGs\_founded &  NA (no relation)\\
  LOC.LOC.containedby
 \end{tabular}
 \caption{Selected relations for relation extraction; 
PPL = people, GOV = governement}
 \label{tab:relations}
\end{table}

For relation extraction, 
we first select the ten most frequent relations 
(plus NA for no relation according to Freebase)
of entity pairs in CF-FIGMENT.
We ensure that the entity pairs have at least one context in CF-FIGMENT.
This results in 
5815, 3054 and 6889 unique entity pairs for train, dev and 
test.\footnote{We only assign those entity pairs to test
  (resp.\ dev, resp.\ train)
for which both constituting entities are in the ET test
(resp.\ dev, resp.\ train) set.}
Dev and test set sizes are 124,462 and
556,847 instances. For the train set, we take
a subsample of 135,171 sentences.
The entity and sentence sets of
CF-FIGMENT were constructed to ensure that entities in the
entity test set do not occur in the sentence train and dev
sets; that is,
a sentence was  assigned to the train set only if all
entities it contains
are train entities.\footnotemark[1]

\subsection{Word, Entity and Type Embeddings}
\seclabel{embeddings}
We use 100-dimensional word embeddings to initialize
the input layer of  
ET
and RE. 
Embeddings are kept fixed during
training.  Since we need embeddings for words, entities and
types in the same space, we process ClueWeb+FACC1 (corpus
with entity information) as follows.  For each sentence $s$,
we add two copies: $s$ itself, and a copy in which each
entity is replaced with its notable type, the most important
type according to Freebase.  We process train, dev and test
this way, but do not replace test entities with
their notable type because the types of test entities 
are unknown in our application scenario.  We run word2vec
\cite{mikolov2013efficient} on the resulting corpus to learn
embeddings for words, entities and types. Note that our
application scenario is that we are given an unannotated input corpus and our
system then extracts entity types and relations from this
input corpus to enhance the KB.

\subsection{Entity Typing Experiments}
\textbf{Entity context setup.}
We use a window size of 5 on each side of the
entity mentions. 
Following \newcite{figment15}, 
we replace other entities occurring in the context with
their Freebase notable type mapped to FIGER. 

\textbf{Models.}
\newcite{figment15} applied a
multi-layer perceptron (\textbf{MLP}) architecture
to create context representations.
Therefore, we use an MLP baseline to compute
the context representation $\vec{\phi}({c})$.
The input to the MLP model is a concatenation of
context word embeddings. As an alternative to MLP, we also
train a \textbf{CNN}  (see \secref{model})
to compute context representations.
We run experiments with
MLP and CNN, each trained with standard distant supervision
and with  MIML.

\textbf{EntEmb and FIGMENT baselines.}
Following \newcite{figment15},
we  also
learn entity embeddings
and classify those embeddings to types, i.e.,
instead of distant supervision, we classify entities based on aggregated information
represented in entity embeddings.
An MLP with one hidden layer is used as classifier.
We call that model EntEmb. 
We join the results
of EntEmb with our best model (line 13 in \tabref{micro}),
similar to the joint model (FIGMENT) in \newcite{figment15}.

We use the same \textbf{evaluation measures} as
\newcite{ling2012fine}, \newcite{figment15} and \newcite{neelakantan2015inferring}
for 
entity typing:
precision at 1 ($P@1$), which is the accuracy of picking
the most confident type for each entity,
micro average $F_1$ of all entity-type assignments 
and  mean average precision (MAP) 
over types. 
We could make assignment decisions based on the
standard criterion 
$p>\theta$, $\theta=0.5$, but we found that
tuning
$\theta$ improves results. 
For each probabilistic
classifier and each type, we set $\theta$ to the value that
maximizes performance on dev.

\textbf{Results.}
\tabref{micro} shows
results for $P@1$, micro $F_1$ and MAP.
For $F_1$, we report separate results for all,
head (frequency higher than
100) and tail (frequency less than 5) entities.

\begin{table}[tb]
\begin{center}
\scriptsize

\begin{tabular}{@{\hspace{0.0cm}}r@{\hspace{0.05cm}}l|lllll}
&& $P@1$ & $F_1$  & $F_1$ & $F_1$ &  MAP\\
&& all  & all & head  &  tail &   \\
\hline
\hline
1&MLP
      & 74.3 & 69.1 & 74.8 & 52.5 & 42.1\\ 
2&MLP+MIML-MAX
      &74.7  & 59.2 & 50.7 & 46.8 &41.3 \\
3&MLP+MIML-AVG
      & 77.2 & 70.6 & 74.9 & 56.2 &45.0 \\
4&MLP+MIML-MAX-AVG
      & 75.2 & 71.2 & 76.4 & 56.0 &47.1 \\
5&MLP+MIML-ATT
      & 81.0 & 72.0 & 76.9 & 59.1 &$48.8$ \\
\hline
6&CNN
      & 78.4 & 72.2 & 77.3 & 56.3 &$47.6$\\
7&CNN+MIML-MAX
      & 78.6 & 62.2 & 53.5 & 49.7 & $46.6$\\
8&CNN+MIML-AVG
      & 80.8 & 73.5 & 77.7 & 59.2 & $50.4$\\
9&CNN+MIML-MAX-AVG
      & 79.9 & 74.3 & 79.2 & 59.8 & $53.3$\\
10&CNN+MIML-ATT
      & 83.4 & 75.1 & 79.4 & 62.2 & $55.2$\\

\hline
11&EntEmb
     & 80.8 & 73.3 & 79.9  & 57.4 & 56.6\\
\hline
12&FIGMENT
    & 81.6  & 74.3  & 80.3  & 60.1 & 57.0 \\
13&CNN+MIML-ATT+EntEmb
   & \textbf{85.4} & \textbf{78.2} & \textbf{83.3} & \textbf{66.2} & \textbf{64.8} \\

\end{tabular}

\caption{$P@1$, Micro $F_1$ for all, head and tail entities and MAP
  results for entity typing. }
\tablabel{micro}
\end{center}
\end{table}

\textbf{Discussion.}
The 
improvement of CNN (6) compared to
MLP (1) is not surprising considering the effectiveness of
CNNs in finding position independent local features, compared
to the flat representation of MLP.  
Lines 2-5 and 7-10 show the results of different MIML algorithms for
MLP and CNN, respectively.
Considering micro F1 for all entities as the most importance measure,
the trend is similar in both MLP and CNN for MIML methods:
ATT $>$ MAX-AVG $>$ AVG $>$ MAX.

MAX is worse than even basic distant supervised models,
especially for micro F1. 
MAX predictions are based on only one context of each entity (for each type),
and the results suggest that this is harmful for entity typing.
This is in contradiction with the previous results in RE (cf. \newcite{Zeng15emnlp}) and suggests that 
there might be a significant difference between expressing 
types of entities and relations between them in text.
Related to this, MAX-AVG which averages the type probabilities at prediction time 
improves MAX by a large margin.
Averaging the context probabilities seems to be a way
to smooth the entity type probabilities. 
MAX-AVG models are also better than the corresponding models with AVG
that train and predict with averaging.
This is due to the fact that AVG gives equal weights 
to all context probabilities both in training and prediction.
ATT uses weighted
contexts in both training and prediction
and that is probably the reason for its effectiveness
over all other MIML algorithms.
Overall, using attention (ATT) significantly improves 
the results of both MLP and CNN models.

CNN+MIML-ATT (10) performs comparable to EntEmb (11),
with better micro F1 on all and tail entities and 
worse MAP and micro F1 on head entities.
These two models have different properties, e.g., MIML is also able to type each mention
of entities while EntEmb works only for corpus-level typing
of entities. (See \newcite{figment15} for more 
differences)
It is important to note that MIML
can also be applied to any entity typing architecture or model
that is trained by distant supervision. Due to the lack of
large annotated corpora, distant supervision is currently 
the only viable approach to fine-grained entity typing;
thus, our demonstration of the effectiveness of MIML is an
important finding for entity typing.

Joining the results of CNN+MIML-ATT with Ent\-Emb (line 13)
gives large improvements over each of the single models.
It is also consistently better (by more than 3\% in all
measures) than our baseline FIGMENT (12),
which is basically MLP+EntEmb.
This improvement is achieved by using CNN instead of MLP for
context representation and integrating MIML-ATT.
EntEmb is improved by \newcite{figment17} by using entity names.
We leave the integration of that model to future work.

\begin{figure}
\includegraphics[width=.45\textwidth]{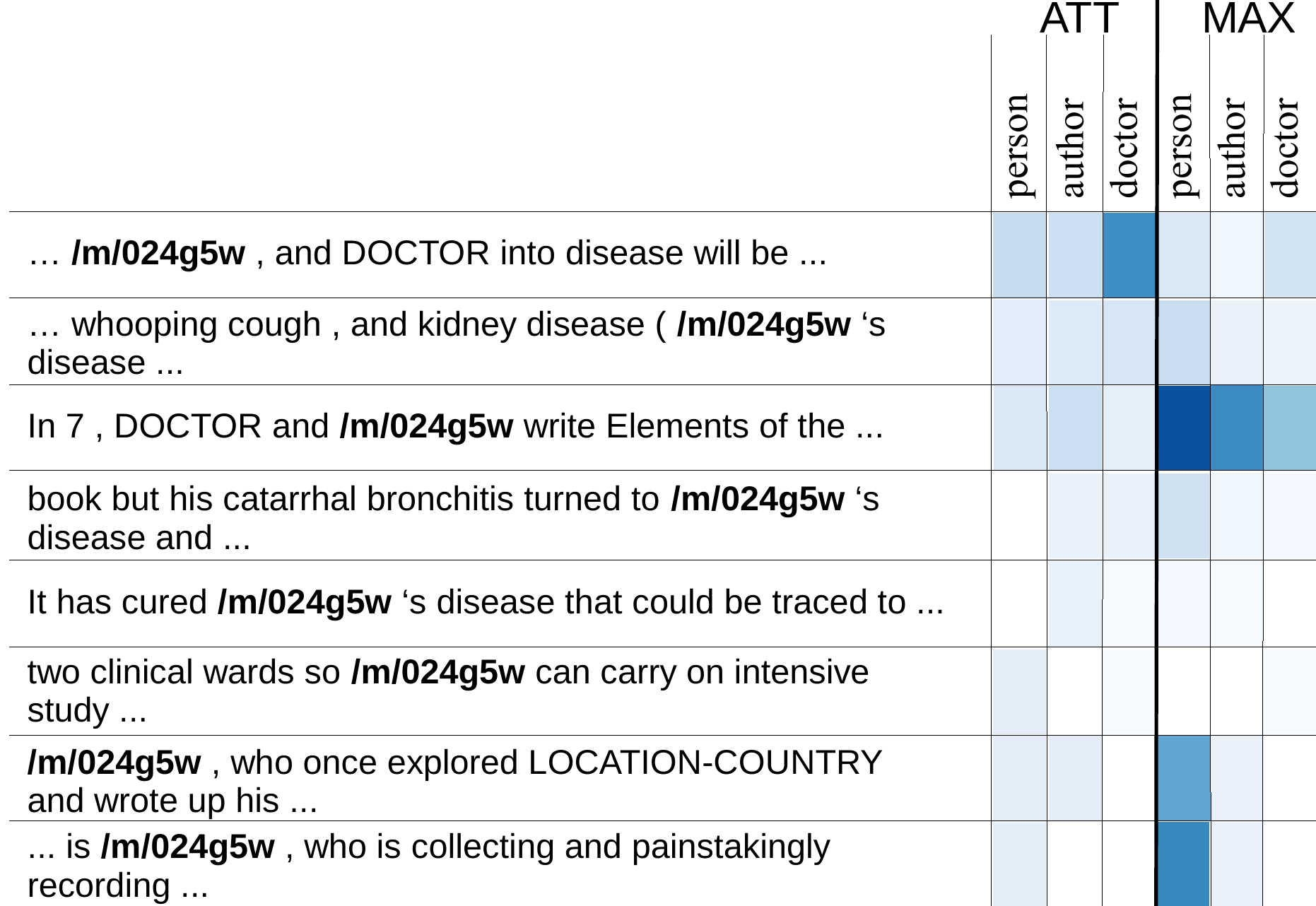}
\caption{MIML-ATT and MIML-MAX scores for the example entity /m/024g5w.
  }
\figlabel{fig:analysis}
\end{figure}

\textbf{Example.}
To show the behavior of MIML-MAX and MIML-ATT, 
we extract the scores that each method 
assigns to the labels for each context. A comparison 
for the example entity ``Richard Bright'' (MID: /m/024g5w) who is a \textsc{person}, \textsc{doctor}
and \textsc{author} is shown in \figref{fig:analysis}. 
Note that the weights from
MIML-ATT (\eqref{alpha}) sum to 1 for each label because of the applied softmax function while the
scores from MIML-MAX (\eqref{cm}) do not.
For both methods, the scores for the type \textsc{person} are more
equally distributed than for the other types which makes sense
since the entity has the \textsc{person} characteristics in every sentence.
For the other types, both models seem to be influenced by other
entities occurring in the context (e.g., an occurrence with another
\textsc{doctor} could indicate that the entity is also a \textsc{doctor}) but also
by trigger words such as ``write'' or ``book'' for the type \textsc{author} or
``disease'' for the type \textsc{doctor}.

\subsection{Relation Extraction Experiments}
\seclabel{reExp}
\textbf{Models.}
In our experiments, we compare two state-of-the-art
RE architectures: piecewise CNN \cite{Zeng15emnlp}
and contextwise CNN  \cite{adel2016}. We use 
the publicly available implementation
for the piecewise CNN \cite{pcnns16url} 
and our own implementation for the contextwise CNN.
Both CNNs represent the input words with embeddings and 
split the contexts based on the
positions of the relation arguments. The context\-wise CNN 
splits the input before convolution, the piecewise CNN 
after convolution. 
Also, while the piecewise CNN applies a softmax layer directly after pooling,
the context\-wise CNN feeds the pooling results into a fully-connected
hidden layer first.
For both models, we use MI learning to mitigate the noise
from distant supervision.

\textbf{Results.}
The precision recall (PR) curves in
\figref{relationExtraction} show that the contextwise CNN outperforms
the piecewise CNN on our RE dataset. We also
compare them to a baseline model 
that does not learn context features but
uses only the embeddings of the relation arguments
as an input
and feeds them into an MLP
(similar
to the EntEmb baseline for ET).
The results confirm that the context features
which the CNNs extract are very important, not only
for ET but also for RE.
Note that the PR curves are 
calculated on the corpus level and not 
on the sentence-level, i.e.,
after aggregating the predictions for
each entity pair. 
Following \newcite{ritter13tacl}, we compute the
area $A$ under the PR curves which supports this 
trend (EntEmb: $A = 0.34$, piecewise CNN: $A = 0.48$,
contextwise CNN: $A = 0.50$).

\textbf{Pipeline vs.\ joint training.}
Since the contextwise CNN
outperforms the piecewise CNN,
we use the contextwise CNN 
for integrating entity types.
Figure \ref{fig:REwithET} shows
that the performance
on the RE dataset increases
when we integrate entity type information into the CNN.
The main trend of the PR curves 
and the areas under them shows the following
order of model performances: 
JOINT-TRAIN $>$ WEIGHTED $>$ PREDICTED-HIDDEN
$>$ BINARY-HIDDEN $>$ BINARY. 

  \begin{figure}
\centering
   \includegraphics[width=.5\textwidth]{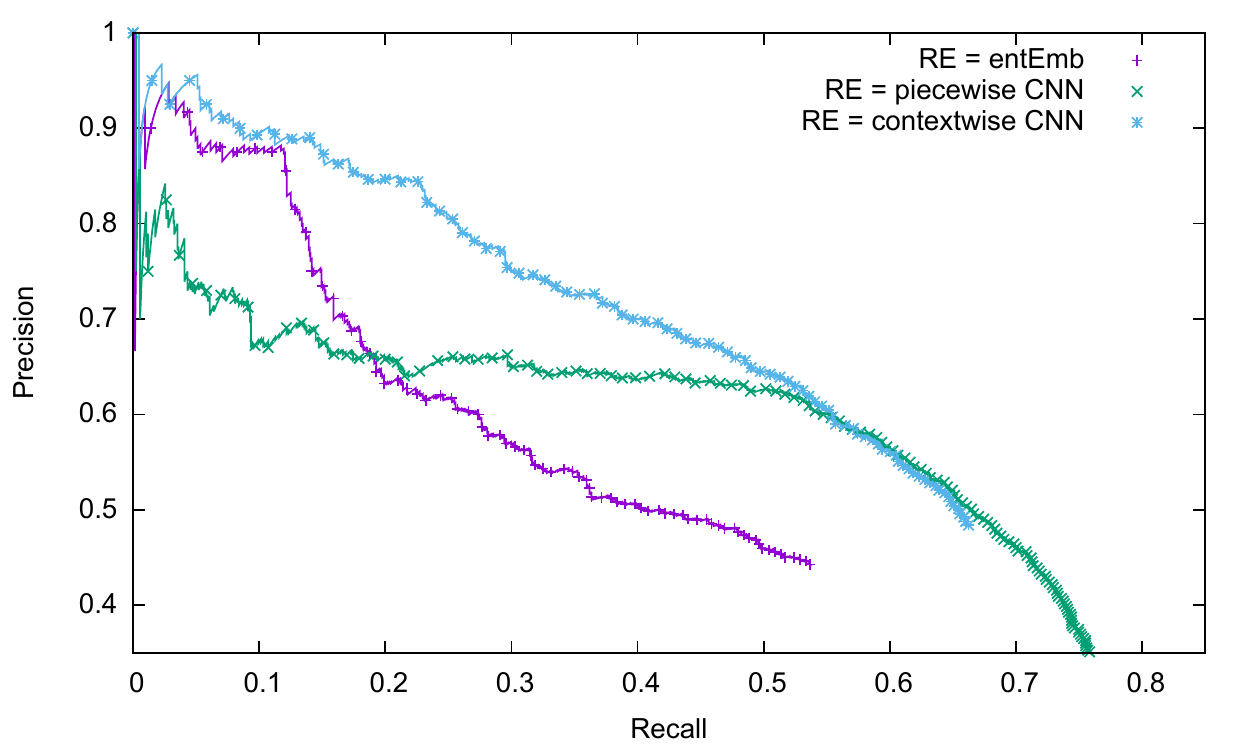}
   \caption{PR curves: relation extraction models}
   \figlabel{relationExtraction}
  \end{figure}
  
    \begin{figure}
\centering
   \includegraphics[width=.5\textwidth]{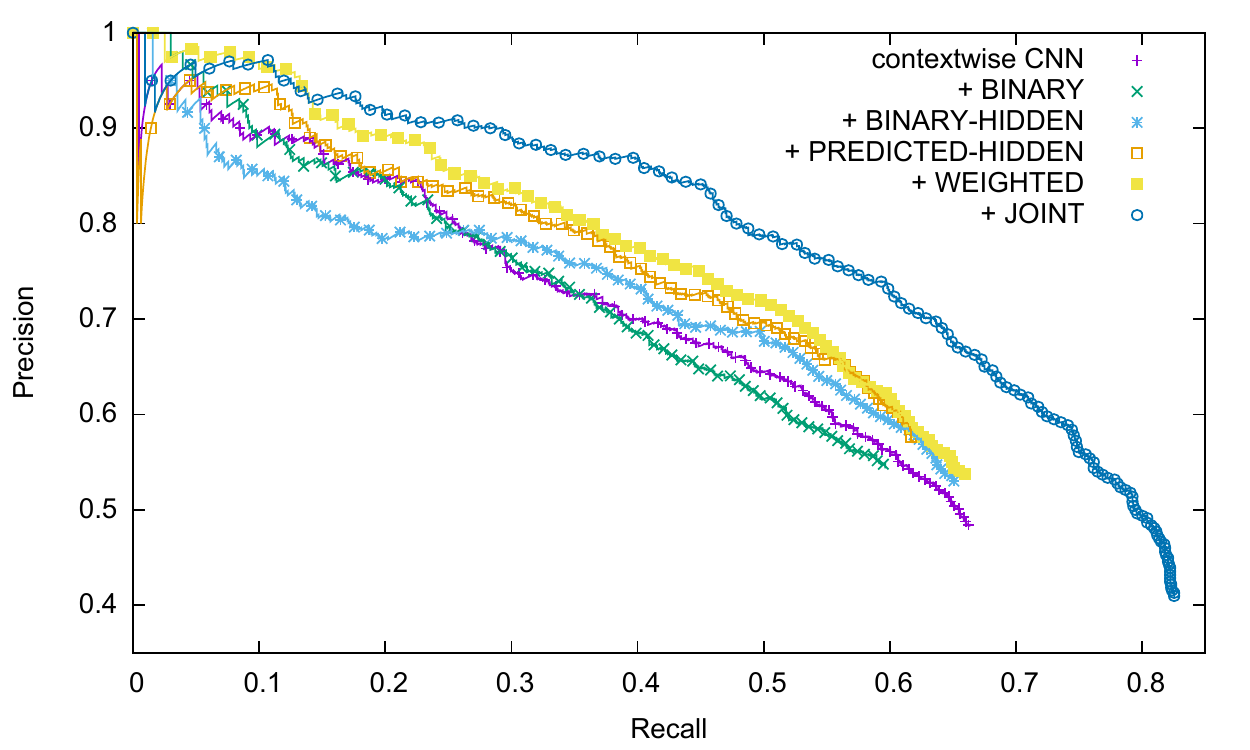}
   \caption{PR curves: type-aware relation extraction models}
   \label{fig:REwithET}
  \end{figure}
  
\textbf{Discussion.}
The better performance of our approaches of integrating
type predictions into the contextwise CNN (PREDICTED-HIDDEN, WEIGHTED) 
compared to
baseline type integrations (BINARY, BINARY-HIDDEN) shows that 
probabilistic predictions of
an entity typing system can be a valuable resource
for RE. 
With binary types, it is not
possible to tell whether one of the selected types
had a higher probability than another or whether a
type whose binary value is 0 just barely missed the threshold.
Probabilistic representations
preserve this
information. Thus, using probabilistic representations, the RE system
can compensate for noise in ET predictions.

WEIGHTED with access to the distributional type embeddings 
learned from the corpus works better than all other pipeline models.
This shows that our type embeddings can be valuable for RE.
JOINT-TRAIN performs better than all pipeline models, even though 
the ET part in the pipelines is trained on 
more data. The area
of JOINT-TRAIN under the PR curve is $A = 0.66$.
A plausible reason is the mutual dependencies of those two tasks
which a joint model can better learn than a pipeline 
model.
We can also relate it to better noise mitigation 
of jointed ET, compared to isolated models.\footnote{On 
the joint dataset, joint training 
improves MAP for entity typing by about 20\% compared to the
best isolated model.}

\textbf{Analysis of joint training.}
In this paragraph, we investigate the joint training
in more detail. 
In particular, we evaluate different variants
of it by combining relation extraction with
other entity typing approaches: EntEmb and FIGMENT.
For joint training with ET-EntEmb, we do not use the context
for predicting the types of the relation arguments but only their embeddings.
Then, we feed those embeddings into an MLP which computes a representation
that we use for the type prediction. This corresponds to
the EntEmb model presented in \tabref{micro} (line 11).
For joint training with ET-FIGMENT, we compute two different cost functions
for entity typing: one for typing based on entity embeddings (see ET-EntEmb above)
and one for typing based on an MLP context model.
This does not correspond exactly to
the FIGMENT model from \tabref{micro} (line 12) which combines an
entity embedding and MLP context model as a postprocessing step but comes close.
In addition to those two baseline ET models, we 
also train a version in which both entity typing and relation
extraction use EntEmb as their only input features. \figref{joint}
shows the PR curves for those models. The curve for the model that
uses only entity embedding features for both entity typing and 
relation extraction
is much worse than the other curves. This emphasizes the importance of
our context model for RE (see also \figref{relationExtraction}), 
also in combination with joint training. 
Similarly, the curve for the model with EntEmb as entity
typing component has more precision variations than the curves for the other
models which use context features for ET. Thus, joint training does not help
per se but it is important which models are trained together.
The areas under the PR curves show the following model trends:
\enote{ha}{I will put this in an enote for now - I think we don't need the
exact $A$ values in the paper

joint with FIGMENT for ET ($A = 0.667$) $>$ joint 
orig ($A = 0.662$) $>$ joint with EntEmb for 
ET ($A = 0.642$) $>$ joint with EntEmb 
for RE+ET ($A = 0.35$)}
joint with ET-FIGMENT $\approx$ joint as in Figure 1 $>$ joint with
ET-EntEmb $>$ joint with ET-EntEmb and RE-EntEmb.

\begin{figure}
 \includegraphics[width=.50\textwidth]{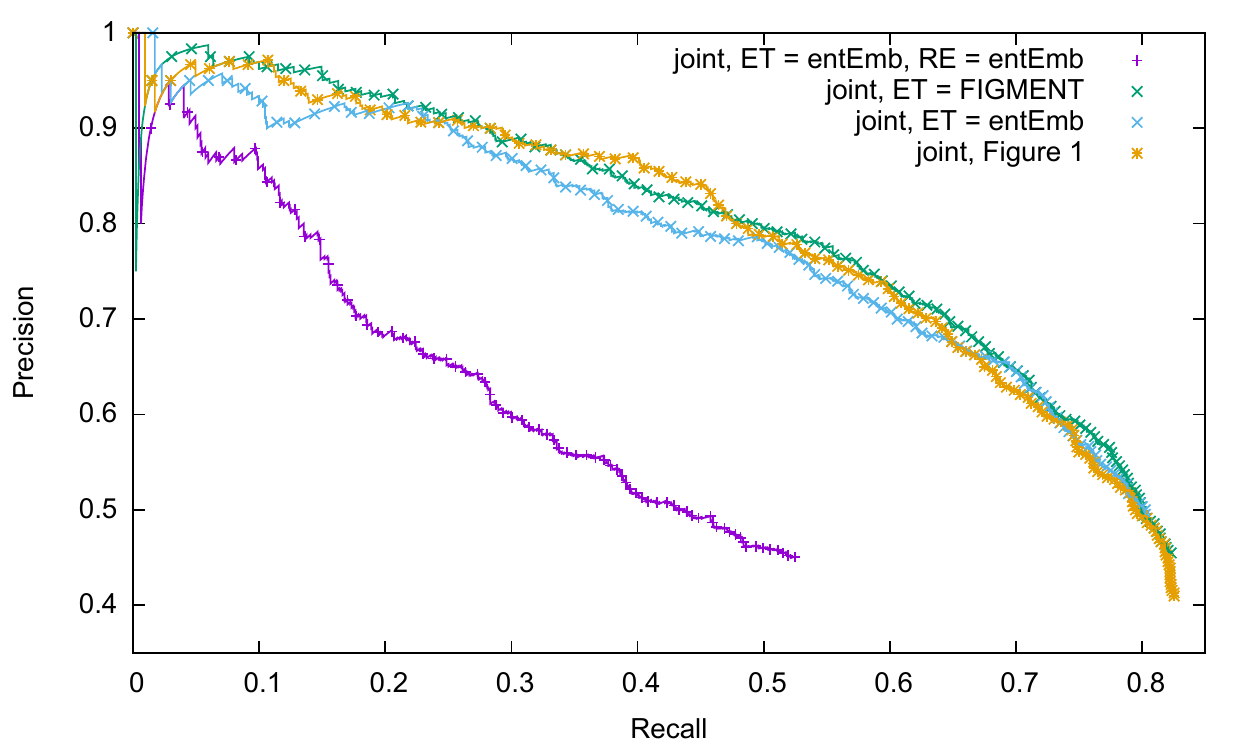}
 \caption{Variants of joint training}
 \figlabel{joint}
\end{figure}

\textbf{Most improved relations.}
To identify which relations are improved the most when entity
types are integrated,
we compare the relation specific $F_1$ scores of CNN,
CNN+WEIGHTED and CNN+JOINT-TRAIN.
With WEIGHTED, the relations PPL.deceased\_PER.place\_of\_death
and LOC.LOC.containedby are
improved the most (from 36.13 to 53.73 and 49.04 to 64.19 $F_1$, resp.).
JOINT-TRAIN has the most positive impact on PPL.deceased\_PER.place\_of\_death, 
ORG.ORG.place\_founded and GOV.GOV\_agen\-cy.jurisdiction 
(from 36.13 to 67.10, 42.38 to 58.51 and 62.26 to 70.41 resp.).

\section{Conclusion}
In this paper, we addressed different types of noise in two
information extraction tasks: entity typing and relation extraction.
We presented the first multi-instance multi-label methods for
entity typing and showed that
it helped to alleviate the noise from distant supervised labels.
This is an important contribution because 
most of the current fine-grained entity typing systems 
are trained by distant supervision.
Our best model sets a new state of the art in corpus-level
entity typing.
For relation extraction, we mitigated noise from using
predicted entity types as features. 
We compared 
different pipeline approaches with each other and 
with our proposed joint type-relation extraction model.
We observed that
using 
type probabilities is more robust 
than binary predictions of types,
and joint training gives the best results.

\section*{Acknowledgments}
This work was supported by DFG (SCHU2246/8-2) and by
a Google European Doctoral
Fellowship granted to Heike Adel.

\bibliography{ref}
\bibliographystyle{eacl2017}

\end{document}